\documentclass[conference]{IEEEtran}
\usepackage{graphicx}

\begin{document}
% \title{Cardiac Magnetic Resonance Fingerprinting\\ Inference from Wi-Fi Channel State Information\\ using Machine Learning}
\title{Automated Schizophrenia Detection from Handwriting Samples via Transfer Learning Convolutional Neural Networks
\vspace{1cm}}


\author{
\IEEEauthorblockN{Rafael Castro}
\IEEEauthorblockA{
Hackley School\\
Tarrytown, NY\\
rcastro@students.hackleyschool.org}
\and
\IEEEauthorblockN{Ishaan Patel}
\IEEEauthorblockA{Emory University\\
Atlanta, GA\\
ishaan.patel@emory.edu}
\and
\IEEEauthorblockN{Tarun Patanjali}
\IEEEauthorblockA{University of Texas at Austin\\
Austin, TX\\
tarun.patanjali@utexas.edu}
\and
\IEEEauthorblockN{Priya Iyer}
\IEEEauthorblockA{University of Michigan\\
Ann Arbor, MI\\
priyai@umich.edu}}

\maketitle

\begin{abstract}
Schizophrenia is a globally prevalent psychiatric disorder that severely impairs daily life. Schizophrenia is caused by dopamine imbalances in the fronto-striatal pathways of the brain, which influences fine motor control in the cerebellum. This leads to abnormalities in handwriting. The goal of this study was to develop an accurate, objective, and accessible computational method to be able to distinguish schizophrenic handwriting samples from non-schizophrenic handwriting samples. To achieve this, data from Crespo et al. (2019) was used, which contains images of handwriting samples from schizophrenic and non-schizophrenic patients. The data was preprocessed and augmented to produce a more robust model that can recognize different types of handwriting. The data was used to train several different convolutional neural networks, and the model with the base architecture of InceptionV3 performed the best, differentiating between the two types of image with a 92\% accuracy rate. To make this model accessible, a secure website was developed for medical professionals to use for their patients. Such a result suggests that handwriting analysis through computational models holds promise as a non-invasive and objective method for clinicians to diagnose and monitor schizophrenia.
\end{abstract}

\IEEEpeerreviewmaketitle

\section{Introduction}

\subsection{Schizophrenia}
Schizophrenia (SCZ) is a psychiatric disorder characterized by hallucinations, delusions, and disordered thinking [1]. As of 2024, SCZ is estimated to affect about 24 million people by the World Health Organization (WHO) [3], and has an estimated economic burden of \$150 billion in the US, which is caused by unemployment, productivity loss, and direct health care costs [4]. However, it is estimated that around 33\% of all people with SCZ are currently undiagnosed, and therefore do not receive care for their condition [10]. Furthermore, current survey-based psychiatric examinations have a limited number of questions, and cannot always capture the full mental state of the individual taking the examination [9]. It is therefore imperative that more efficient, objective, and accurate examinations are developed to diagnose schizophrenia so these patients can receive professional help for their condition.

\subsection{Motor Abnormalities}
SCZ is caused by overactive fronto-striatal pathways in the brain which cause the patient to associate their thoughts indiscriminately with reality [2]. The striatum and the fronto-striatal are part of the cortical-striatal circuit, which influences the fine motor control in the cerebellum [5]. Therefore, abnormalities such as dopamine imbalances in the fronto-striatal pathways can influence motor control, which is linked to handwriting deformities [2]. Furthermore, these handwriting deformities have been shown to worsen with condition [2], giving psychiatrists and other medical professionals an objective tool to measure the severity of one's psychiatric condition.

\begin{figure}[htp]
    \centering
    \includegraphics[width=7cm]{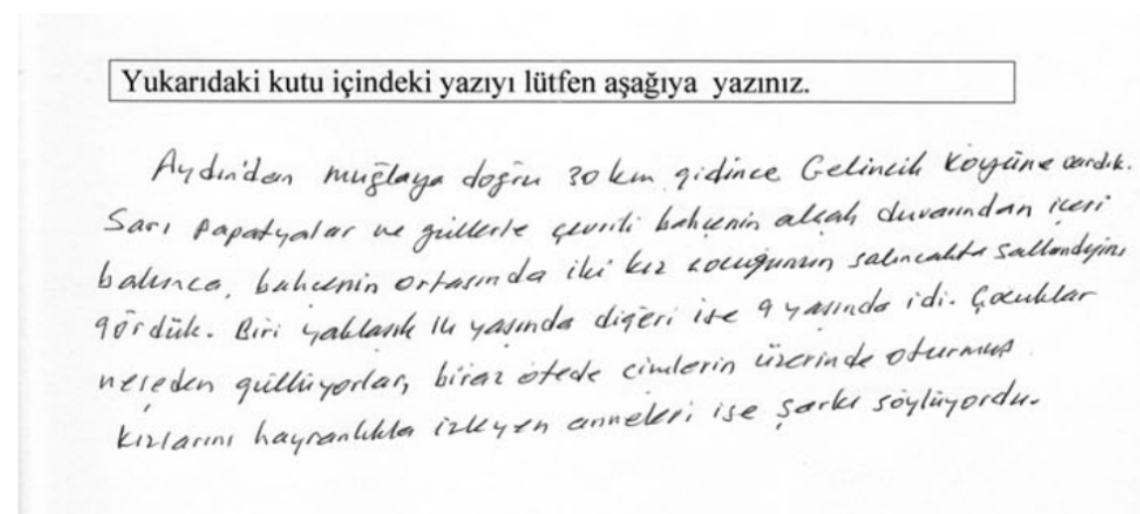}
    \caption{An image of handwriting taken from a non-schizophrenic patient.}
    \label{fig:galaxy}
\end{figure}

\begin{figure}[htp]
    \centering
    \includegraphics[width=7cm]{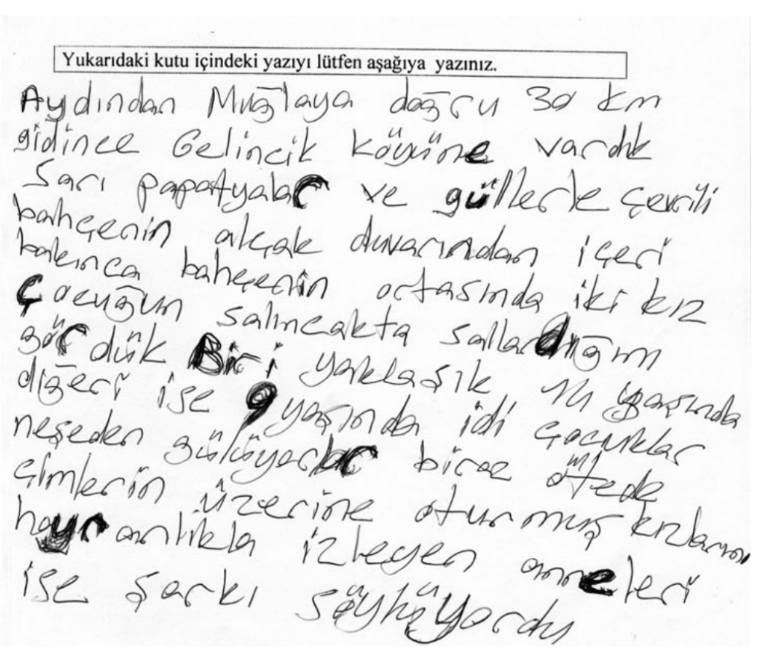}
    \caption{An image of handwriting taken from a schizophrenic patient.}
    \label{fig:galaxy}
\end{figure}

\subsection{Literature Review}

Crespo et al. (2019) [7] sought to understand the difference in handwriting between SCZ patients and healthy controls. They collected handwriting data from images, and converted it to tabular data. The researchers then performed t-tests and U-tests on measures such as length and spaces in handwriting. Overall, motor abnormalities were shown to be significantly elevated in SCZ patients (p $<$ 0.01).

Komür et al. (2019) [6] sought to understand the difference in handwriting in SCZ patients before and after treatment. collected data from handwritten images, and also converted it to tabular data. The researchers found a statistically significant difference in handwriting samples from SCZ patients before treatment and after treatment. The study showed a decrease (p $<$ 0.05) in height and width of individual letters in schizophrenic patients, and a change (p $<$ 0.05) in line inclination, showing that motor abnormalities decreased with treatment.

Caliguri et al. (2022) [2] created a linear regression model that was able to account for 83\% of the variability in medication dose for SCZ patients using only three factors derived from handwriting kinematics. This model was able to differentiate between SCZ and non-SCZ handwriting much more accurately than medical professionals, showing the importance of developing computational methods to differentiate between SCZ and non-SCZ handwriting. This suggests that handwriting kinematics are a biomarker of striatal dopamine dysfunction, which is a chemical imbalance in the striatum that can affect motor control in the cerebellum.

\subsection{Statement Of Purpose}

This study hopes to accomplish three goals. The first is to develop a computational model to automatically analyze handwriting features without involvement from a data scientist or medical professional. The second goal is to make the model accessible, as this technology should be readily available to everyone. The third goal is to output an objective score that represents one's psychiatric condition, as it is imperative to bring objectivity to psychiatric data science.

\section{Methods}

\subsection{Data Collection}

The data for this project was obtained from Crespo et al. (2019) [7], which are images of four loops which both SCZ and non-SCZ patients were instructed to follow with a digital pen. In Crespo et al., data was collected for the experimental group that had been diagnosed with SCZ from individuals at the Mental Health Day Unit at the University St. Agustin Hospital in Spain. The data from the control group were from the University of Jaén and an adult school of Jaén, which are both also in Spain.

\begin{figure}[htp]
    \centering
    \includegraphics[width=8cm]{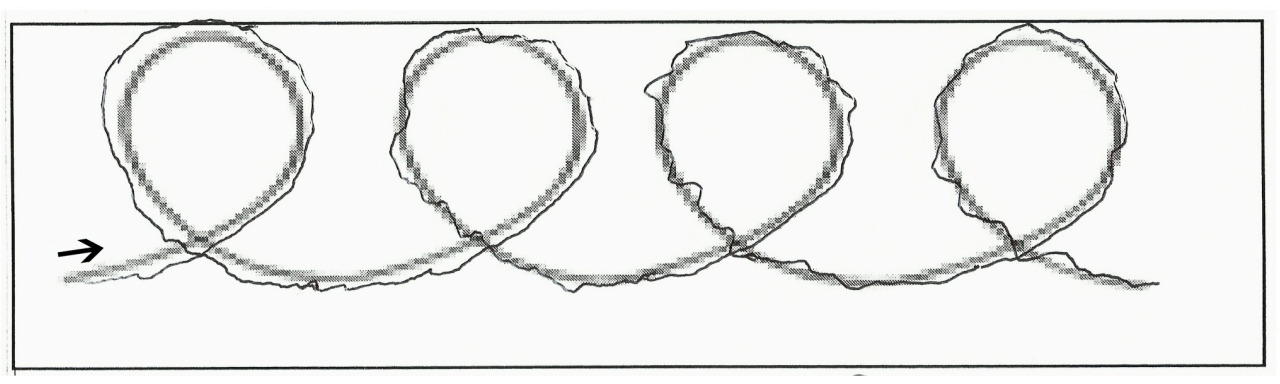}
    \caption{The participants' view of data collection. Participants followed the loops on an A4 paper affixed to a digital tablet with a digital pen, and the Ductus software was used to collect data.}
    \label{fig:galaxy}
\end{figure}

\subsection{Data Pre-processing}

The data from Crespo et al. (2019) [7] was provided with graph lines and a label on the axes that said "Posición". Furthermore, there was padding around the graphs that was inconsistent with each image. Four pre-processing steps were taken to make the data as uniform as possible, which produces a more reliable and accurate model.

The first pre-processing step taken was cropping the image. This was done manually with Adobe Photoshop, and removed any excess padding, the graph lines, and the title. After this, the data was centered to ensure the data was uniformly cropped, and that any discrepancies that may have occured by cropping the images manually were dealt with. Next, the data received padding to make sure that each image was the exact same size. This was done using the \texttt{pillow} package in Python. Finally, the Laplacian of Gaussian (LoG), an advanced edge detection filter that measures the 2nd spatial derivative of an image [18], was applied to the data. The LoG filter is defined as the Laplace operator applied to the Gaussian filter, as shown by the equation $\Delta(I * G) = I * \Delta G$, where I is the image, G is the Gaussian kernel, and $*$ is the convolution operator.

\begin{figure}[htp]
    \centering
    \includegraphics[width=8cm]{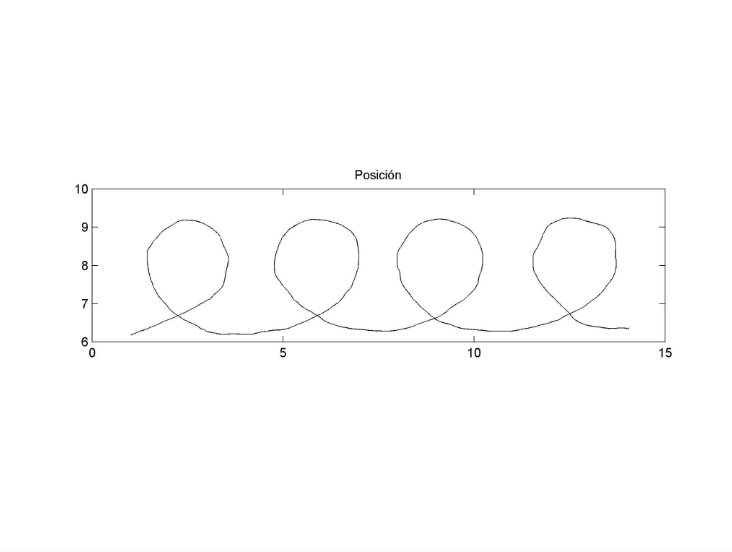}
    \caption{The image data that was received from the researchers.}
    \label{fig:galaxy}
\end{figure}

\begin{figure}[htp]
    \centering
    \includegraphics[width=8cm]{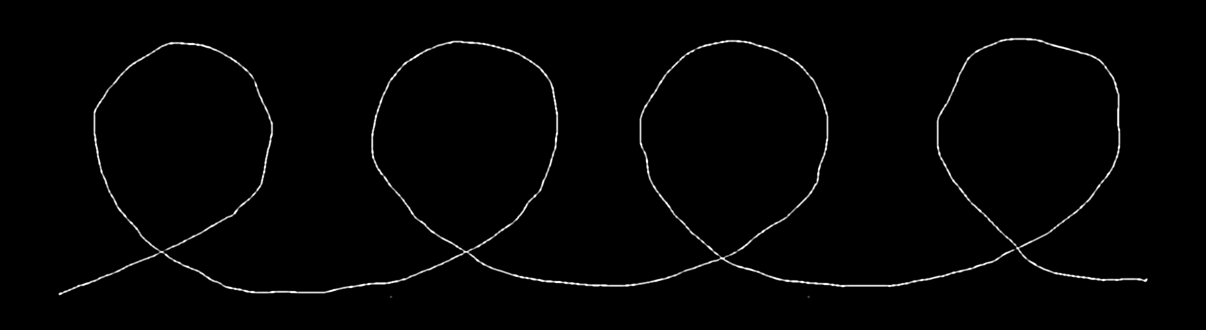}
    \caption{The image data after undergoing all four steps of pre-processing: cropping, centering, padding, and the LoG filter.}
    \label{fig:galaxy}
\end{figure}

\subsection{Data Augmentation}

Two types of data augmentation were performed to increase the robustness of the model and mimic other types of handwriting. The first type of data augmentation was shearing, which randomly changes the perception angle of the image. This mimics slanted handwriting, and preserves the structure of the loops while generalizing the data to other, more slanted types of handwriting. The second type of data augmentation performed was a horizontal flip, which also preserves the content of the loops while mimicking other styles of handwriting. With these augmentation steps, the size of the dataset increased to 720 images. For a machine learning study, it is important to note that even with the new data, this is a small sample size.

\subsection{Model Selection}

The data analyzed was in the form of images, so convolutional neural networks (CNNs) were used to build the model. Three different CNN models were tested against each other on the dataset. To test these models, 80\% of the data was used for training, 10\% of the data was used for validation, and 10\% of the data was used for testing. The three models tested were the Inception V3 model [11], the Efficient Net B3 model [12], and a custom CNN attached to a feed-forward neural network. Inception V3 has been shown to work well in the past with classifying handwriting samples as a transfer learning model [19] [20]. EfficientNet B3 is also a very highly regarded CNN for transfer learning. We also wanted to include our own, smaller, CNN and compare it to the larger transfer learning models to see if a less computationally complex model would perform on a similar level as the larger transfer learning models. For all of these models, the Adam optimizer, a learning rate of 0.0001, and the cross-entropy loss function were employed.

\begin{figure}[htp]
    \centering
    \includegraphics[width=8cm]{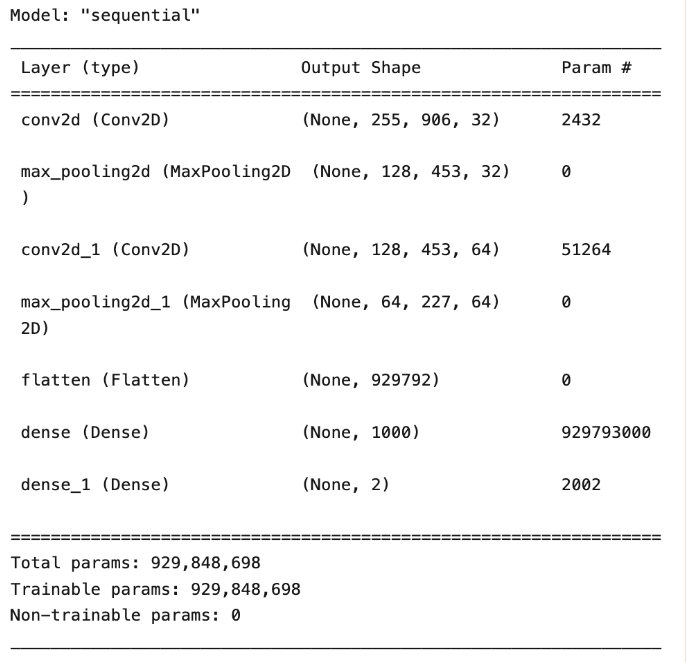}
    \caption{The model architecture of the custom CNN model used.}
    \label{fig:galaxy}
\end{figure}

\section{Results}

\subsection{Model Performance}

After training each model, we evaluated its performance using the accuracy metric, and the Inception V3 model performed the best, with an accuracy of 93\%. Figure 8 shows the confusion matrix for the InceptionV3 model, with the "control" label denoting healthy patients and "patient" label denoting SCZ patients.

\begin{figure}
\begin{center}
\begin{tabular}{ |p{3cm}||p{3cm}|  }
 \hline
 \multicolumn{2}{|c|}{Model Performance} \\
 \hline
 Model Name&Accuracy\\
 \hline
 InceptionV3&91.7\%\\
  Custom CNN&72.2\%\\
 EfficientNetB3&63.9\%\\
 \hline
\end{tabular}
\end{center}
\caption{The table showing the performance of the three CNN models used. InceptionV3, a widely used transfer learning model, performed the best.}
\end{figure}

\begin{figure}[htp]
    \centering
    \includegraphics[width=8cm]{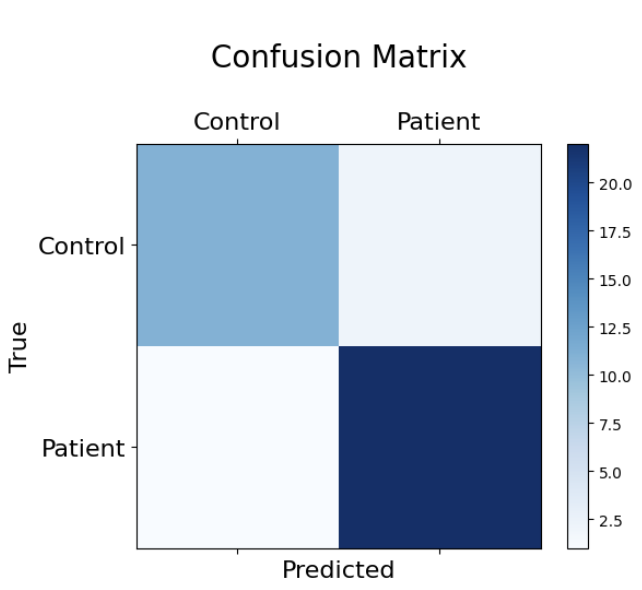}
    \caption{The confusion matrix showing the performance of the InceptionV3 model. The model performed exceptionally well, with an accuracy of about 92\%.}
    \label{fig:galaxy}
\end{figure}

\section{Discussion}

\subsection{Accessibility}

In order to make this technology available to anyone who would need it, an internal website was developed using \texttt{react.js}, GitHub Pages, and Amazon Web Services (AWS). The website, which is intended for the use of psychiatrists or medical professionals, hosts the model online using AWS, and allows the user to upload a file with a handwriting sample. The website then processes the image and feeds it into the model, and outputs a number from 0 to 1 representing the probability that the handwriting sample comes from a SCZ patient.

\subsection{Limitations}

The models in this study were trained on a very small dataset (n = 240), as there is not much data that has been collected on handwriting samples from people suffering from schizophrenia. This was addressed by augmenting the data collected, but the models developed in this paper still may not be robust or able to be generalized to populations outside of Spain. Furthermore, due to regional handwriting variation (the phenomena that handwriting is similar in one country or region) [17], it is likely that the model picked up on specific aspects of Spanish handwriting. To test this, it is imperative to collect more data globally and retrain the models. It is also important to mention the ethical implications of using machine learning for these types of studies, such as algorithmic biases that cannot be easily detected, as neural networks are a type of "black-box" model.

\subsection{Potential Application}

Attempts to have human analysis of features in handwriting for specific conditions such as schizophrenia have been largely unsuccessful. However, researchers have been able to construct mathematical models, such as the one in Caliguri et al (2009) [2], that can classify disorders such as schizophrenia from handwriting samples. This work introduces a more sophisticated type of model to this problem, CNNs, that have a similar accuracy to the linear regression used in Caliguri et al (2009), and shows the promise of transfer learning being used in this field.

Furthermore, this study shows the promise of computational analysis of handwriting samples as a tool for doctors and other medical professionals to use in screening for schizophrenia. Collecting handwriting data that is similar to what was analyzed in this study is quick and non-invasive, compared to length psychiatric examinations. While handwriting analysis is not yet as accurate as other methods, such as questionnaires, this field is largely unexplored and holds promise for the future.

\section{Conclusion}
This study achieved three goals. Firstly, a highly accurate model was developed for automatic analysis of handwriting samples for signs of motor abnormalities caused by SCZ, and achieved an accuracy of 92\%. Secondly, the model was made accessible through a website designed for psychiatrists and other medical professionals, which hosts the model securely through Amazon Web Services. Finally, this model represents an objective way to represent the condition of a SCZ patient, as handwriting kinematics have been shown to be a biomarker of condition severity and striatal dopamine dysfunction [2] [5].

\subsection{Future Work}
In the future, ways to expand this would be to collect more data to produce a more robust model. Furthermore, this study could be expanded to use natural language processing techniques to analyze the textual content of handwriting samples for linguistic features of schizophrenia, such as auditory verbal hallucinations [13] [14] [15] [16].

\section*{Acknowledgment}
The authors would like to thank Christian Cardozo for his guidance with the project and Dr. Yasmina Crespo for readily making the data from her study available.


\begin{thebibliography}{1}

\bibitem{IEEEhowto:kopka}
Schak, KM, and JG Leung. “Schizophrenia.” Mayo Clinic, Mayo Clinic, 7 Jan. 2020, www.mayoclinic.org/diseases-conditions/schizophrenia/symptoms-causes/syc-20354443. 

\bibitem{IEEEhowto:kopka} Caligiuri, Michael P et al. “Handwriting movement analyses for monitoring drug-induced motor side effects in schizophrenia patients treated with risperidone.” Human movement science vol. 28,5 (2009): 633-42. doi:10.1016/j.humov.2009.07.007


\bibitem{IEEEhowto:kopka} “Schizophrenia.” World Health Organization, World Health Organization, 10 Jan. 2022, www.who.int/news-room/fact-sheets/detail/schizophrenia. 


\bibitem{IEEEhowto:kopka} Cloutier, Martin et al. “The Economic Burden of Schizophrenia in the United States in 2013.” The Journal of clinical psychiatry vol. 77,6 (2016): 764-71. doi:10.4088/JCP.15m10278


\bibitem{IEEEhowto:kopka} Hayhow, Bradleigh D et al. “The neuropsychiatry of hyperkinetic movement disorders: insights from neuroimaging into the neural circuit bases of dysfunction.” Tremor and other hyperkinetic movements (New York, N.Y.) vol. 3 tre-03-175-4242-1. 26 Aug. 2013, doi:10.7916/D8SN07PK


\bibitem{IEEEhowto:kopka} Kömür, İlhami et al. “Differences in Handwritings of Schizophrenia Patients and Examination of the Change after Treatment.” Journal of forensic sciences vol. 60,6 (2015): 1613-9. doi:10.1111/1556-4029.12858


\bibitem{IEEEhowto:kopka} Crespo Y, Ibañez A, Soriano MF, Iglesias S, Aznarte JI (2019) Handwriting movements for assessment of motor symptoms in schizophrenia spectrum disorders and bipolar disorder. PLoS ONE 14(3): e0213657. https://doi.org/10.1371/journal.pone.0213657


\bibitem{IEEEhowto:kopka} Michael P Caligiuri, Peter J Weiden, Anna Legedza, Sergey Yagoda, Amy Claxton, Handwriting Kinematics in Patients with Schizophrenia Treated with Long-Acting Injectable Atypical Antipsychotics: Results From the ALPINE Study, Schizophrenia Bulletin Open, Volume 3, Issue 1, January 2022, sgac018, https://doi.org/10.1093/schizbullopen/sgac018  


\bibitem{IEEEhowto:kopka} Gornale, S., Kumar, S., Siddalingappa, R., \& Hiremath, P. S. (2022). Survey on Handwritten Signature Biometric Data Analysis for Assessment of Neurological Disorder using Machine Learning Techniques. Transactions on Engineering and Computing Sciences, 10(2), 27–60. https://doi.org/10.14738/tmlai.102.12210


\bibitem{IEEEhowto:kopka} “Could You Have Schizophrenia and Not Know It?” Could You Have Schizophrenia and Not Know It?: Allied Psychiatry \& Mental Health: Psychiatrists, Allied Psychiatry \& Mental Health. Accessed 2 Feb. 2024. 


\bibitem{IEEEhowto:kopka} Szegedy, Christian, et al. "Rethinking the inception architecture for computer vision." Proceedings of the IEEE conference on computer vision and pattern recognition. 2016.


\bibitem Tan, Mingxing, and Quoc Le. "Efficientnet: Rethinking model scaling for convolutional neural networks." International conference on machine learning. PMLR, 2019.

\bibitem Tang, S.X., Kriz, R., Cho, S. et al. Natural language processing methods are sensitive to sub-clinical linguistic differences in schizophrenia spectrum disorders. npj Schizophr 7, 25 (2021). https://doi.org/10.1038/s41537-021-00154-3

\bibitem{IEEEhowto:kopka} Buck, Benjamin, and David L Penn. “Lexical Characteristics of Emotional Narratives in Schizophrenia: Relationships With Symptoms, Functioning, and Social Cognition.” The Journal of nervous and mental disease vol. 203,9 (2015): 702-8. doi:10.1097/NMD.0000000000000354

\bibitem{IEEEhowto:kopka} Chang, X., Zhao, W., Kang, J. et al. Language abnormalities in schizophrenia: binding core symptoms through contemporary empirical evidence. Schizophr 8, 95 (2022). https://doi.org/10.1038/s41537-022-00308-x

\bibitem{IEEEhowto:kopka} Bae YJ, Shim M, Lee WH. Schizophrenia Detection Using Machine Learning Approach from Social Media Content. Sensors. 2021; 21(17):5924. https://doi.org/10.3390/s21175924

\bibitem{IEEEhowto:kopka} Bhana, Yusuf. “Handwriting ‘Accents’: How We Write Reveals Our Cultural Identity.” Toppan Digital Language, 9 Nov. 2018, toppandigital.com/us/blog-usa/handwriting-accents-penmanship-can-betray-language-identity/. 

\bibitem{IEEEhowto:kopka} Dhar, Rajdeep \& Gupta, Radheshyam \& Baishnab, Krishna. (2014). An analysis of CANNY and LAPLACIAN of GAUSSIAN image filters in regard to evaluating retinal image. 1-6. 10.1109/ICGCCEE.2014.6922270.

\bibitem{IEEEhowto:kopka} Tallapragada, V.V.S., Alivelu Manga, N., Nagabhushanam, M.V., Venkatanaresh, M. (2022). Greek Handwritten Character Recognition Using Inception V3. In: Somani, A.K., Mundra, A., Doss, R., Bhattacharya, S. (eds) Smart Systems: Innovations in Computing. Smart Innovation, Systems and Technologies, vol 235. Springer, Singapore. https://doi.org/10.1007/978-981-16-2877-1\_23

\bibitem{IEEEhowto:kopka} R. Gayathri and R. Babitha Lincy. 2022. Transfer learning based handwritten character recognition of tamil script using inception-V3 Model. J. Intell. Fuzzy Syst. 42, 6 (2022), 6091–6102. https://doi.org/10.3233/JIFS-212378

\end{thebibliography}
\end{document}